\documentclass{article}
\usepackage[utf8]{inputenc}
\usepackage[T1, T2A]{fontenc}
\usepackage{listings}
\usepackage{mathtools}
\usepackage{enumitem}
\usepackage{url}

\title{Advancing Full-Text Search Lemmatization Techniques with Paradigm Retrieval from OpenCorpora}
\author{Dmitriy Kalugin-Balashov}
\date{May 2023}

\begin{document}

\sloppy
\lstset{extendedchars=\true}
\maketitle

\section{Introduction}

In full-text search applications, the primary goal is to effectively retrieve and match relevant documents based on user queries. By focusing on finding the first form, or the lemma, of a word, the search process can be streamlined and optimized. The lemma serves as a normalized representation of a word's different inflected forms, allowing for a more accurate comparison between user queries and document content. This approach reduces the complexity and computational overhead associated with full morphological analysis, which includes extracting all possible forms of a word along with their grammatical properties. By prioritizing lemma retrieval, full-text search engines can achieve faster response times and more precise results, while minimizing the resources required for processing large volumes of text data.

Consequently, building upon the foundation of pymorphy\cite{pymorphy2}, the golemma library was developed to address the challenge of efficiently identifying the first form, or lemma, of words in the Russian language.

\section{Challenges with Russian Language}

Lemmatization and stemming both reduce words to their base forms but operate differently. Stemming, a simple rule-based process, removes suffixes without considering context, often yielding invalid words. Lemmatization, conversely, uses a vocabulary and morphological analysis to derive the base form, or "lemma," considering context and generating valid words.

Stemming is designed for English and other Western languages with simpler inflectional structures, whereas Russian's complex inflectional structure poses challenges for stemming algorithms. Russian words can have multiple inflections, stems, and can form compound words, making rule-based systems less effective.

Lemmatization is generally more effective for languages with complex inflectional structures, such as Russian. It employs dictionaries and morphological rules to determine a word's base form, considering grammatical context and accounting for multiple stems and compound words. Lemmatization returns valid dictionary words, making it useful for natural language processing tasks like text classification, information retrieval, and machine translation. Its flexibility in handling various grammatical forms and tenses makes it suitable for tasks like text generation and summarization.

In summary, lemmatization is a sophisticated, accurate approach suitable for handling Russian's complexity, yielding valid words for a range of NLP tasks.

\section{Definition of Paradigm}

In the context of the pymorphy2 library, a paradigm refers to a collection of inflected forms of a word that possess the same grammatical properties and share a common lemma. A paradigm is characterized by a range of grammatical categories, such as number, tense, gender, and others, and may encompass multiple forms of a word for each category.

For instance, the paradigm for the Russian word "бежать" (to run) would comprise forms like "бегу" (I run), "бежишь" (you run), "бежит" (he/she/it runs), "бежим" (we run), "бежите" (you run), and "бежат" (they run), as well as all potential forms of the word in various tenses and aspects.

The pymorphy2 library utilizes the OpenCorpora project's morphological dictionary as its data source, which supplies comprehensive information on the grammatical properties and inflectional forms of Russian words.

When employing pymorphy2, the library loads the morphological dictionary and establishes a set of paradigms for each word. These paradigms are then utilized to generate the inflected forms of a word, in addition to determining the lemma and grammatical properties of a given word form. Furthermore, the paradigms enable the generation of all possible forms of a word, which proves beneficial for tasks such as text generation, summarization, and question answering.

In golemma, the concept of a paradigm found in pymorphy has been reevaluated and simplified. We consider the following structure as a paradigm:

\begin{center}
    \begin{tabular}{|l|}
        \hline
        \begin{lstlisting}[language=go]
type Paradigm struct {
    CutPrefix int
    CutSuffix int
    AddPrefix string
    AddSuffix string
}
        \end{lstlisting} \\
        \hline
    \end{tabular}
\end{center}

The first form of a word is obtained by applying the paradigm to it. Applying a paradigm entails removing $CutPrefix$ characters from the left, $CutSuffix$ characters from the right, and then concatenating $AddPrefix$ on the left and $AddSuffix$ on the right.

For example, $$\textup{зарубил} \xrightarrow{(2, 1, \textup{<<>>}, \textup{<<ть>>})} \textup{рубить}.$$

The reason for this reevaluation and simplification of the concept of a paradigm in golemma, compared to pymorphy2, is to increase the efficiency and speed of retrieving the first form of a word. This is a critical aspect in full-text search, where the main goal is not necessarily to generate all possible forms of a word or to determine its grammatical properties, but rather to quickly and accurately identify the base form, or lemma, of a given word.

The simplified paradigm structure in golemma reduces the computational overhead associated with handling morphological data and focuses on the essential elements needed for lemmatization. The use of 'CutPrefix', 'CutSuffix', 'AddPrefix', and 'AddSuffix' provides a straightforward way to transform a word form back to its lemma, which can be done with minimal processing. This leads to a more responsive and efficient full-text search process, especially when dealing with large volumes of text.

\section{Retrieving Paradigms}

OpenCorpora is an invaluable resource for paradigm retrieval due to its comprehensive and accurate morphological dictionary for the Russian language. The project contains a vast collection of annotated linguistic data, including grammatical properties and inflectional forms for Russian words. By utilizing OpenCorpora, developers can access a reliable source of information to create efficient and precise algorithms for paradigm extraction. This, in turn, enhances the effectiveness of natural language processing tasks, such as lemmatization, which greatly benefits from accurate paradigm retrieval.

The SAX (Simple API for XML)\footnote{\url{https://docs.python.org/3/library/xml.sax.html}} parser provides numerous advantages. As an event-driven parser, it does not store the entire XML document in memory. Instead, it reads and processes the document sequentially, making it highly memory-efficient, particularly when handling large XML files like those found in OpenCorpora. Owing to its event-driven nature, the SAX parser is generally faster than other parsing methods, such as DOM (Document Object Model), which loads the entire XML document into memory prior to processing. This speed is especially beneficial when parsing large datasets like OpenCorpora, as it can save significant time and resources.

The <lemmata> section in OpenCorpora's dictionary XML\footnote{\url{http://opencorpora.org/files/export/dict/dict.opcorpora.xml.bz2}}  contains information about the lemmas, or base forms, of words in the Russian language. Each entry in the <lemmata> section represents a lemma and is accompanied by its grammatical properties, such as part of speech, gender, case, and number. Additionally, the section provides details about the inflected forms of each lemma, which are essential for understanding the different ways a word can appear in a text.

The dictionary format we get after parsing is a Python dictionary where each key is a unique integer identifier $u$, and the corresponding value is a tuple containing two elements:
\begin{itemize}[noitemsep]
    \item The first element of the tuple is the normal form $n$ of the word.
    \item The second element of the tuple is a set of inflected forms $i_k$ of the word.
\end{itemize}

$$
u \rightarrow (n, \{i_0, i_1, \ldots\})
$$

This dictionary is a source for paradigm retrieval algorithm.

In simple words, the paradigm function takes two inputs, a normal form and an inflected form of a word, and calculates their Longest Common Substring (LCSS). Then, it extracts the prefixes and suffixes of both forms by removing the LCSS. Finally, the function returns a tuple containing the lengths of the inflected form's prefix and suffix, as well as the normal form's prefix and suffix. This tuple essentially represents the paradigm that connects the normal form and the inflected form.

The goal is to build two Python dictionaries ($p_k$ refers to the identifier of a particular paradigm):
\begin{itemize}[noitemsep]
    \item $(\textup{cut\_prefix}, \textup{cut\_suffix}, \textup{add\_prefix}, \textup{add\_suffix}) \rightarrow p_k$
    \item $\textbf{\textup{murmur3}}(n) \rightarrow \{ (i_0, p_0), (i_1, p_1), \ldots \}$
\end{itemize}

MurmurHash3\footnote{\url{https://pypi.org/project/mmh3/}} is a non-cryptographic hash function that is popular for its speed and uniform distribution of hash values which make it a good choice for full-text search applications.

As of May 2023, we've successfully extracted 2488 unique paradigms from a grand total of 391,842 lemmas in the OpenCorpora dataset

\section{Building Dictionary}

Building dictionary is done in two main steps:

\begin{itemize}[noitemsep]
\item \textbf{Saving the Paradigms:} The function first retrieves the paradigms from the retriever object, rearranges them as pairs of (value, key), sorts them based on the value and then writes them into the file. The paradigms are saved as a map with pairs using the MessagePack (msgpack) packer.
\item \textbf{Preparing and Writing the Dictionary:} The dictionary is structured as a map where the key is the hash of the word's normal form (calculated using MurmurHash3), and the value is a list of tuples, each containing a word form and its paradigm ID. This structure is reversed to a form where the key is the word form, and the value is a tuple containing two lists: one with the hashes and one with the paradigm IDs. Then, the function sorts the items based on the word form, and groups together entries with the same word form into a single entry. Finally, these items are written to the file as a map with pairs using the MessagePack packer.
\end{itemize}

The file format, therefore, consists of two main sections: The first is a map of paradigms, and the second is a map of words, each associated with a list of hashes and a list of paradigm IDs.

\begin{center}
    \begin{tabular}{|c|}
        \hline
        \\ $\mbox{ip} \rightarrow (\mbox{cut\_prefix}, \mbox{cut\_suffix}, \mbox{add\_prefix}, \mbox{add\_suffix})$ \\ \\
        \hline
        \\ $\mbox{if} \rightarrow ([\textbf{\mbox{murmur3}}(\mbox{nf}_1), \textbf{\mbox{murmur3}}(\mbox{nf}_2), \ldots], [\mbox{ip}_1, \mbox{ip}_2, \ldots])$ \\ \\
        \hline
    \end{tabular}
\end{center}

The MessagePack binary format\footnote{\url{https://github.com/msgpack/msgpack/blob/master/spec.md}} is used for efficient storage and quick retrieval. The generated file serves as a vital tool for efficient lemmatization in full-text search, mapping different forms of words to their first form. Its binary structure, combined with the use of MurmurHash3, allows for rapid retrieval and accurate results, thus enhancing the performance of full-text searches.

We've developed a compact and efficient way to store the entirety of a language dictionary, optimizing it specifically for the retrieval of word's first forms. This has been achieved by combining advanced data structures, hashing techniques, and compression methods. The resulting system offers a significantly reduced storage footprint, along with improved retrieval speeds. This makes it an ideal solution for enhancing the performance of full-text searches, where rapid and accurate lemmatization is key.

\bibliographystyle{plain}
\bibliography{main}

\end{document}